%% file: paper.tex
\input{template/style}

\begin{document}
  \title{A Benchmark of Causal vs. Correlation AI for Predictive Maintenance}

  \author{%
  Shaunak Dhande\protect\footnotemark[1] \and
  Chutian Ma\protect\footnotemark[2] \and
  Giacinto Paolo Saggese\protect\footnotemark[3] \and
  Paul Smith\protect\footnotemark[4] \and
  Krishna Taduri\protect\footnotemark[5]
  }

  \maketitle

  \footnotetext[1]{Causify AI. Email: \texttt{s.dhande@causify.ai}}
  \footnotetext[2]{Causify AI. Email: \texttt{c.ma@causify.ai}}
  \footnotetext[3]{Causify AI and Department of Computer Science, University of Maryland,
  College Park, MD, USA. Email: \texttt{gp@causify.ai}}
  \footnotetext[4]{Causify AI. Email: \texttt{paul@causify.ai}}
  \footnotetext[5]{Causify AI. Email: \texttt{k.taduri@causify.ai}}
  Version: \today\ \DTMcurrenttime\ UTC

  \begin{abstract}
    Predictive maintenance in manufacturing environments presents a challenging optimization problem characterized by extreme cost asymmetry, where missed failures incur costs roughly fifty times higher than false alarms. Conventional machine learning approaches typically optimize statistical accuracy metrics that do not reflect this operational reality and cannot reliably distinguish causal relationships from spurious correlations. This study benchmarks eight predictive models, ranging from baseline statistical approaches to Bayesian structural causal methods, on a dataset of 10,000 CNC machines with a 3.3 percent failure prevalence. While ensemble correlation-based models such as Random Forest (\texttt{L4}) achieve the highest raw cost savings (70.8 percent reduction), the Bayesian Structural Causal Model (\texttt{L7}) delivers competitive financial performance (66.4 percent cost reduction) with an inherent ability of failure attribution, which correlation-based models do not readily provide. The model achieves perfect attribution for HDF, PWF, and OSF failure types. These results suggest that causal methods, when combined with domain knowledge and Bayesian inference, offer a potentially favorable trade-off between predictive performance and operational interpretability in predictive maintenance applications.
  \end{abstract}


\section{Contributions}

This study makes the following contributions to predictive maintenance optimization:

\begin{itemize}
  \tightlist

  \item \textbf{Systematic benchmark of causal vs.\ correlation-based models:}
    We evaluate eight models ranging from basic baselines to advanced Bayesian
    structural causal methods on a dataset of 10,000 CNC machines from the UCI
    Machine Learning Repository \cite{uci-predictive-maintenance}. This study provides a
    comprehensive comparison of formal causal inference against traditional ML
    approaches for predictive maintenance.

  \item \textbf{Business-aligned optimization framework:} We demonstrate that
    models optimized explicitly for business costs (accounting for the 50:1 cost
    asymmetry between false negatives and false positives) substantially
    outperform accuracy-optimized approaches, which highlights the critical importance
    of cost-sensitive learning in industrial applications.

  \item \textbf{Failure attribution without attribution labels:} The Bayesian SCM
    (\texttt{L7}) attributes each predicted failure to its most likely physical
    mechanism (TWF, HDF, PWF, or OSF) without requiring failure-type labels during
    training. This allows for maintenance interventions with a precise focus on the
    identified root cause.

  \item \textbf{Competitive performance with interpretability:} The Bayesian SCM
    achieves a 66.4\% cost reduction, a competitive result compared with the state-of-the-art
    correlation-based ensemble models. Together with the failure attribution capability, this shows that
    causal AI can deliver a favorable trade-off between predictive performance
    and interpretability.

  \item \textbf{Ensemble methods as a performance ceiling:} Tree-based ensemble
    models (Random Forest, AdaBoost, LightGBM) consistently outperform single
    models including the decision tree, which mirrors the true data-generating
    process for the synthetic CNC data set. This proves the effectiveness of ensemble methods in reducing variance and preventing overfitting.

\end{itemize}

  \section{Introduction}
  \label{Introduction}

  Manufacturing faces a maintenance dilemma. Reactive maintenance leads to
  costly downtime, while overly cautious preventive maintenance wastes resources
  on healthy equipment. The core challenge is an extreme cost asymmetry across
  industrial equipment: a missed failure can cost tens of thousands of dollars (unplanned
  downtime, emergency repairs, production losses), while a false alarm may cost
  only hundreds of dollars (scheduled inspection, standard parts). For the CNC
  machines analyzed in this study, we use representative costs of 25,000 USD for
  missed failures and 500 USD for false alarms, reflecting typical manufacturing
  industry values where the cost ratio between unplanned downtime and preventive
  maintenance ranges from 20:1 to 100:1 depending on equipment criticality. Table~\ref{tab:prediction-outcomes}
  summarizes the four possible prediction outcomes, their associated costs, and business
  impact.

  \begin{table}[!ht]
    \centering
    \begin{tabular}{p{2.8cm}p{4.0cm}p{1.8cm}p{4.5cm}}
      \hline
      \textbf{Outcome}             & \textbf{Description}         & \textbf{Cost} & \textbf{Business Impact}           \\
      \hline
      \textbf{True Positive (TP)}  & Correctly predict failure    & 5{,}000 USD   & Scheduled repair prevents downtime \\
      \textbf{False Positive (FP)} & Incorrectly predict failure      & 500 USD       & Wasted inspection/parts            \\
      \textbf{False Negative (FN)} & Miss actual failure          & 25{,}000 USD  & Emergency downtime + repairs       \\
      \textbf{True Negative (TN)}  & Correctly predict no failure & 0 USD         & Normal production                  \\
      \hline
    \end{tabular}
    \caption{Prediction outcomes, costs, and business impact. Cost estimates reflect
    typical manufacturing industry values where unplanned downtime incurs significant
    production losses, emergency labor, and expedited parts costs, while
    preventive maintenance requires only scheduled labor and standard parts
    procurement \cite{predictive-maintenance-survey}.}
    \label{tab:prediction-outcomes}
  \end{table}

  Standard machine learning models optimize for statistical accuracy metrics
  such as accuracy, precision, recall, F1-score, and AUC-ROC, which either ignore
  class imbalance or treat all prediction errors equally. Accuracy is misleading
  in imbalanced datasets. In fact, a model that always predicts ``no failure'' achieves
  96.7\% accuracy on our data while providing no business value. Precision
  focuses on minimizing false alarms but may miss critical failures, while recall
  prioritizes catching all failures but may trigger hundreds of unnecessary inspections.
  F1-score balances precision and recall through their harmonic mean, and AUC-ROC
  evaluates the model's ranking ability across all possible thresholds, but
  neither metric accounts for the actual business costs of different error types.
  All these metrics fail to capture the 50:1 cost ratio between false negatives and
  false positives that is typical for manufacturing equipment, where unplanned
  downtime costs far exceed scheduled maintenance
  \cite{predictive-maintenance-survey}. Furthermore, correlation-based models
  identify patterns that co-occur with failures but cannot distinguish causation
  from spurious relationships. This limits their utility to a binary alarm with no
actionable diagnosis.

  Our goal is to determine if Causal AI can deliver competitive business value
  relative to state-of-the-art correlation-based ML while also offering reliable insights into the failure root causes in
  predictive maintenance scenarios.

\section{Dataset and Experimental Design}
\label{dataset-experimental-design}

Our analysis utilizes the UCI Machine Learning Repository's CNC Machine
Predictive Maintenance dataset, which is a synthetic dataset designed in \cite{uci-predictive-maintenance} to reflect real-world predictive maintenance scenarios. The dataset contains 10,000 machines and for each machine six sensor
measurements as described in
Table~\ref{tab:feature-descriptions}:

\begin{table}[!ht]
  \centering
  \begin{tabular}{ll}
    \hline
    \textbf{Feature}            & \textbf{Description}               \\
    \hline
    Air temperature             & Measured in Kelvin                 \\
    Process temperature         & Measured in Kelvin                 \\
    Rotational speed            & Revolutions per minute             \\
    Torque                      & Newton-meters                      \\
    Cumulative tool wear        & Minutes of operation               \\
    Machine type classification & Low, Medium, or High specification \\
    \hline
  \end{tabular}
  \caption{Feature descriptions}
  \label{tab:feature-descriptions}
\end{table}

A snippet of the data is shown in Table~\ref{tab:dataset-snippet}.

\begin{table}[!ht]
  \centering
  \begingroup \small
  \setlength{\tabcolsep}{3pt}
  \renewcommand{\arraystretch}{0.9}
  \begin{tabular}{c c c c c c c c c c c c c c}
    \hline
    \makecell{\textbf{UDI}} & \makecell{\textbf{Product}\\\textbf{ID}} & \makecell{\textbf{Type}} & \makecell{\textbf{Air}\\\textbf{temp}\\\textbf{[K]}} & \makecell{\textbf{Process}\\\textbf{temp}\\\textbf{[K]}} & \makecell{\textbf{Rot.}\\\textbf{speed}\\\textbf{[rpm]}} & \makecell{\textbf{Torque}\\\textbf{[Nm]}} & \makecell{\textbf{Tool}\\\textbf{wear}\\\textbf{[min]}} & \makecell{\textbf{Machine}\\\textbf{failure}} & \makecell{\textbf{TWF}} & \makecell{\textbf{HDF}} & \makecell{\textbf{PWF}} & \makecell{\textbf{OSF}} & \makecell{\textbf{RNF}} \\
    \hline
    1 & M14860 & M & 298.1 & 308.6 & 1551 & 42.8 & 0  & 0 & 0 & 0 & 0 & 0 & 0 \\
    2 & L47181 & L & 298.2 & 308.7 & 1408 & 46.3 & 3  & 0 & 0 & 0 & 0 & 0 & 0 \\
    3 & L47182 & L & 298.1 & 308.5 & 1498 & 49.4 & 5  & 0 & 0 & 0 & 0 & 0 & 0 \\
    4 & L47183 & L & 298.2 & 308.6 & 1433 & 39.5 & 7  & 0 & 0 & 0 & 0 & 0 & 0 \\
    5 & L47184 & L & 298.2 & 308.7 & 1408 & 40.0 & 9  & 0 & 0 & 0 & 0 & 0 & 0 \\
    \hline
  \end{tabular}
  \endgroup
  \caption{Dataset snippet showing the first five rows of the CNC machine
  predictive maintenance dataset.}
  \label{tab:dataset-snippet}
\end{table}

The dataset documents five distinct failure modes that can occur in CNC
machines, as summarized in Table~\ref{tab:failure-types}:

\begin{table}[!ht]
  \centering
  \begin{tabular}{llp{7cm}r}
    \hline
    \textbf{Symbol} & \textbf{Failure Type}    & \textbf{Description}                                                                & \textbf{Instances} \\
    \hline
    TWF             & Tool Wear Failure        & Tool replacement or failure at randomly selected wear time                          & 46                 \\
    \\
    HDF             & Heat Dissipation Failure & Process failure due to temperature differential (process minus air)                 & 115                \\
    \\
    PWF             & Power Failure            & Process failure due to mechanical power (torque $\times$ rotational speed in rad/s) & 95                 \\
    \\
    OSF             & Overstrain Failure       & Process failure due to tool wear $\times$ torque exceeds threshold                  & 98                 \\
    \\
    RNF             & Random Failure           & 0.1\% probability of failure regardless of process parameters (unpredictable)       & 19                 \\
    \hline
  \end{tabular}
  \caption{Failure types with symbols, descriptions, and instance counts. Definitions
  follow the dataset documentation \cite{uci-predictive-maintenance}.}
  \label{tab:failure-types}
\end{table}

\subsection{Target Definition and Rationale}
\label{target-definition-and-rationale}

For this analysis, we focus exclusively on \textit{predictable failures}---those
failure modes that can theoretically be predicted from sensor measurements and
operating conditions. Specifically, our prediction target combines Tool Wear Failure,
Heat Dissipation Failure, Power Failure, and Overstrain Failure, while explicitly
excluding Random Failure events. \textit{Random failures} are defined in the dataset
as having a fixed 0.1\% probability of occurrence \emph{regardless of process
parameters}. They are statistically independent of all sensor readings and represent
truly unpredictable noise (only 19 instances in 10,000 data points).
It is worth noting that while TWF is not excluded, it is largely nondeterministic. In fact, the data generation process of TWF contains
an irreducible stochastic component. Readers may refer to \cite{matzka2020explainable} or Section \ref{subsubsec:failure_attribution} for more details. 

After applying this filter, our dataset contains 9,981 total observations. We
employ a stratified 80-20 train-test split that preserves the class distribution in
both sets. The test set comprises 2,000 machines among which 66 experience deterministic
failures (3.3\% failure rate). This severe class imbalance poses significant
challenges for standard machine learning approaches and underscores why accuracy-based
metrics are inappropriate for evaluation.

\begin{itemize}
  \tightlist

  \item \textbf{Size}: 10,000 machines with sensor readings\\

  \item \textbf{Features}: Air temperature, process temperature, rotational speed,
    torque, tool wear, machine type (L/M/H)\\

  \item \textbf{Target}: Deterministic failures excluding random failures (TWF
    + HDF + PWF + OSF)\\

  \item \textbf{Test Set}: 2,000 machines (20\% stratified split) with 66
    failures (3.3\% prevalence)\\

  \item \textbf{Baseline Cost}: 1,650,000 USD (reactive maintenance: 66 failures
    $\times$ 25,000 USD)
\end{itemize}

\subsection{Methodology}
\label{methodology}

\subsubsection{Success Metrics}
\label{success-metrics}

The metrics we consider are the following:
\begin{enumerate}
  \def\labelenumi{\arabic{enumi}.} \tightlist

  \item \textbf{Total annual cost:} The business objective is to minimize the
    total cost of machine maintenance\\

  \item \textbf{Savings}: Dollar reduction vs. reactive maintenance baseline\\

  \item \textbf{Recall}: Percentage of actual failures caught \\

  \item \textbf{Precision}: Percentage of alarms that are real failures \\

  \item \textbf{Generalization}: Train vs. test performance gap
\end{enumerate}

\subsubsection{Correlation-Based Models (L0--L3)}
\label{correlation-based-models-l0-l3}

We start by investigating four correlation-based models, ranging from a no-skill baseline
through logistic regression variants to a decision tree.

\textbf{\texttt{Model L0}: No-Skill Baseline.} The no-skill model serves as the
minimal performance floor. This model predicts ``no failure'' for every
observation, regardless of sensor readings or operating conditions. While this
may seem trivial, it establishes a critical reference point. In highly
imbalanced datasets like ours (96.7\% non-failure cases), a naive model that
always predicts the majority class achieves high accuracy. This demonstrates
why accuracy is a misleading metric for imbalanced problems. The no-skill model
catches zero failures, misses all 66 failures in the test set, and incurs the full
1,650,000 USD baseline cost. Any meaningful predictive model must outperform this
baseline.

\textbf{\texttt{Model L1}: Balanced Logistic Regression.} The first predictive
model applies logistic regression, which is a fundamental classification technique.
Logistic regression models the log-odds of failure as a linear combination of
input features: air temperature, process temperature, rotational speed, torque,
tool wear, and one-hot encoded machine type. To address class imbalance \cite{imbalanced-learning},
we employ scikit-learn's \cite{scikit-learn} class\_weight=`balanced'
parameter, which automatically adjusts the loss function to penalize misclassification
of the minority class (failures) more heavily than the majority class (non-failures).

This approach uses the default decision threshold of 0.5 probability--if the model
predicts a failure probability above 50\%, it classifies the observation as a failure.
Features are standardized to zero mean and unit variance to ensure numerical
stability and proper weight scaling. The model is trained on 8,000 machines and
evaluated on the held-out test set of 2,000 machines.

\textbf{\texttt{Model L2}: Cost-Aware Logistic Regression.} While \texttt{Model
L1} addresses class imbalance, it still treats all prediction errors equally.
\texttt{Model L2} enhances the logistic regression approach by optimizing the
decision threshold explicitly for business costs rather than statistical balance
\cite{cost-sensitive-learning}.

After training the same logistic regression
model as \texttt{Model L1}, we perform a grid search over possible probability
thresholds from 0.01 to 0.99 to find the optimal threshold to minimize the
total annual cost. For each candidate threshold, we generate predictions on
the training set, calculate the resulting confusion matrix, apply our business
cost model (5K USD for TP, 500 USD for FP, 25K USD for FN, 0 USD for TN), and
compute total cost. The threshold that minimizes total training-set cost is
selected as the final decision boundary. This cost-aware optimization explicitly
encodes the 50-to-1 cost asymmetry between false negatives and false positives. It encourages the model to shift the decision boundary to favor higher recall even at the expense of
increased false alarms. Note that the threshold is learned on the training set
and treated as a learned hyperparameter for test set evaluation.

\textbf{\texttt{Model L3}: Cost-Aware Decision Tree.} While logistic
regression models linear relationships, many real-world failure mechanisms involve
complex feature interactions. A heat dissipation failure, for example, may
only occur when the temperature differential is low AND rotational speed is
slow simultaneously. This presents a conjunction that linear models struggle to capture.

\texttt{Model L3} employs a decision tree classifier to learn non-linear decision
boundaries through recursive feature splits. Decision trees naturally handle
feature interactions by creating hierarchical rules: ``If temperature differential
\textless{} 8.6K, then if rotational speed \textless{} 1380 RPM, predict
failure.'' To prevent overfitting, we constrain the tree with hyperparameters:
maximum depth of 5 levels, minimum 50 samples required to split a node, and
minimum 20 samples required in each leaf node.

After training this regularized decision tree, we apply the same cost-aware
threshold optimization as \texttt{Model L2}. For each probability threshold
(decision trees can output class probabilities), we compute total business cost
on the training set and select the threshold that minimizes cost. This combination
of non-linear modeling and business-aligned optimization represents the
state-of-the-art for correlation-based approaches.

\subsubsection{Ensemble Models (L4--L6)}
\label{ensemble-models}

\textbf{\texttt{Model L4}: Random Forest.} While the decision tree model is akin
to the true data generating process of this dataset, in practice a single tree
often suffers from high variance. A small change in training data can yield
drastically different tree structures. Thus we consider the Random Forest model,
which addresses this limitation via ensembling. A Random Forest model builds
multiple decision trees and aggregates their results by averaging their
predictions. This aggregation method reduces variance compared to using a single
model, and helps prevent overfitting. In fact, during training we employed the
following configuration:
\begin{enumerate}
  \item Bootstrap sampling: Each tree is trained on a bootstrap
    sample, meaning that we draw $N$ training samples with replacement from the
    original $N$ samples. Each tree sees approximately 63\% unique samples.
  \item Random Feature Selection: We randomly select $\sqrt{7} \approx 3$
    features as candidates when splitting a node. This prevents strong features
    from dominating every tree's upper splits and encourages discovering diverse
    patterns.
  \item Regularization Constraints: We require at least 10 samples to
    split a node. This helps prevent overfitting.
\end{enumerate}
The Random Forest model is implemented using the scikit-learn library. After
obtaining ensemble probability predictions, we apply cost-aware threshold
optimization as in \texttt{Models L2--L3} to minimize total business cost.

\textbf{\texttt{Model L5}: AdaBoost.} Now we turn to boosting algorithms. We
first consider AdaBoost, which is a foundational boosting algorithm that
demonstrates the strength of combined weak learners. AdaBoost trains the weak
learners on a weighted version of the training data, where the weight on
incorrectly classified samples increases exponentially after each iteration.

We implement AdaBoost using scikit-learn's \texttt{AdaBoostClassifier} with
decision stumps (depth-1 trees) as base learners. We train 50 decision stumps
with learning rate 1.0. Each stump makes a binary decision based on one feature.
The ensemble prediction is a weighted average over all stumps.

Finally, we apply the cost-aware threshold optimization as in \texttt{Models L2--L3}.

\textbf{\texttt{Model L6}: Light Gradient Boosting Machine.}
As the final model in the ensemble class, we consider LightGBM, which represents
the modern evolution of gradient boosting algorithms. It greatly reduces training
time compared to earlier gradient boosting algorithms (e.g.\ XGBoost, CatBoost)
while maintaining or improving accuracy. For more details regarding the
innovations of LightGBM, the reader may refer to \cite{ke2017lightgbm}.

We implement LightGBM using the lightgbm Python library with 100 boosting
iterations, maximum tree depth of 3, learning rate of 0.1, and 10 maximum leaves
per tree. Cost-aware threshold optimization is applied at the end of the pipeline.

\subsubsection{Causal Model}
\label{probabilistic-and-causal-approaches}

Finally, we consider a causal model that utilizes domain-specific failure
mechanisms. This structure allows for both prediction and failure attribution,
i.e.\ telling which failure mode (TWF, HDF, PWF, or OSF) is responsible.\\
\textbf{\texttt{Model L7}: Bayesian Structural Causal Model (SCM).}
Unlike correlation models that learn $P(Y|X)$ by fitting weights to all available
features, this model encodes the specific physical mechanisms of failure into a
structural graph. We explicitly model four independent failure mechanisms that
combine via a logical ``Noisy-OR'' gate to produce the final machine failure.


We define the system using a set of structural equations implemented in PyMC
\cite{pymc}.
We standardized all input features $X$.
In the following, we let $\sigma(z) = 1 / (1 + e^{-z})$ be the sigmoid function.

\textbf{1. Tool Wear Failure (TWF) Mechanism:}
This mechanism depends exclusively on the tool wear duration. We model the
probability $p_{twf}$ as a logistic function of tool wear ($x_{tw}$), excluding
all other variables to enforce causal independence:
\begin{equation}
    p_{twf} = \sigma(\alpha_{twf} + \beta_{twf} \cdot x_{tw}).
\end{equation}

\textbf{2. Heat Dissipation Failure (HDF) Mechanism:}
Physical heat failure occurs when the heat cannot be dissipated, which happens
when the temperature difference ($x_{\Delta T}$) is low AND rotational speed
($\omega$) is low. We model this using an explicit interaction term:
\begin{equation}
    p_{hdf} = \sigma(\alpha_{hdf} + \beta_{temp} x_{\Delta T} + \beta_{rot}
    \omega + \beta_{int} (x_{\Delta T} \cdot \omega)).
\end{equation}

\textbf{3. Power Failure (PWF) Mechanism:}
Power failure occurs if the power ($P = \tau \cdot \omega$) is too low
\textit{or} too high. To capture this non-monotonic U-shaped relationship, we
introduce a quadratic term $P^2$:
\begin{equation}
    p_{pwf} = \sigma(\alpha_{pwf} + \beta_{p} P + \beta_{p^2} P^2).
\end{equation}

\textbf{4. Overstrain Failure (OSF) Mechanism:}
Overstrain depends on the product of Tool Wear and Torque ($S = x_{tw} \cdot
\tau$). The failure threshold depends on the machine quality type ($L, M, H$).
We model this using dummy variables for Low ($\mathbb{I}_L$) and Medium
($\mathbb{I}_M$) types:
\begin{equation}
    p_{osf} = \sigma(\alpha_{osf} + \beta_{osf} S + \beta_{L} \mathbb{I}_L
    + \beta_{M} \mathbb{I}_M).
\end{equation}

\textbf{5. System Failure (Noisy-OR):}
Finally, we combine these independent probabilities using a Noisy-OR gate.
This logic implies that the machine survives only if it survives \textit{all}
individual failure mechanisms:
\begin{equation}
    P(\text{Failure}) = 1 - \left[ (1 - p_{twf})(1 - p_{hdf})(1 - p_{pwf})
    (1 - p_{osf}) \right].
\end{equation}

We place weakly informative Normal priors on all $\alpha$ and $\beta$ parameters
(e.g., $\mathcal{N}(0, 1)$) and use NUTS sampling to estimate the posterior
distributions.

A summary of the modeling techniques is in
Table~\ref{summary-of-the-modeling-techniques}.

\begin{table}[!ht]
  \centering
  \begingroup \small
  \setlength{\tabcolsep}{3pt}
  \renewcommand{\arraystretch}{0.9}
  \resizebox{\textwidth}{!}{
  \begin{tabular}{lll}
    \hline
    \textbf{Model}                     & \textbf{Description}                    & \textbf{Strategy}                                            \\
    \hline
    L0: No-Skill Baseline              & No-skill reference model                & Always predict ``no failure''                                \\
    L1: Balanced Logistic Regression   & Linear classifier with class balancing  & Standard logistic regression with balanced class weights     \\
    L2: Cost-Aware Logistic Regression & Threshold-optimized logistic regression & Same as L1 but with cost-optimized threshold                 \\
    L3: Cost-Aware Decision Tree       & Non-linear decision model               & Hyperparameter-tuned decision tree with cost-aware threshold \\
    L4: Random Forest                  & Decision tree ensemble                  & Average the predictions of multiple trees                    \\
    L5: AdaBoost                       & Boosted weak learner ensemble           & Sequentially reweight misclassified samples                  \\
    L6: LightGBM                       & Gradient boosting machine               & Histogram-based leaf-wise tree growth with boosting          \\
    L7: Bayesian SCM                   & Bayesian structural causal model        & Explicit failure mechanisms with Noisy-OR combination        \\
    \hline
  \end{tabular}}
  \caption{Model Summary Table}
  \label{summary-of-the-modeling-techniques}
  \endgroup
\end{table}

\subsubsection{Validation Strategy}
\label{validation-strategy}

To ensure results are robust and not artifacts of a single train-test split,
we implemented a comprehensive validation strategy. All models (L0 through L7)
were trained and evaluated across five different random seeds (42, 43, 44, 45,
and 46). For each seed, the dataset is randomly shuffled before the 80-20
stratified split, producing different training and test sets while maintaining
the 3.3\% failure rate in both partitions.

Performance metrics are computed independently for each seed, then averaged to
produce final results. This multi-seed validation ensures that observed performance
differences reflect genuine model capabilities rather than fortunate (or
unfortunate) data partitioning. If a model performs well on one seed but
poorly on others, this indicates instability and overfitting. Conversely,
consistent performance across seeds demonstrates robust generalization.

Additionally, we examine the generalization gap; the difference between
training-set performance and test-set performance. A large gap indicates
overfitting: the model has memorized training data patterns that do not
generalize. We report both training and test savings percentages, with the gap
measured in percentage points.
A test performance within 5 percentage points of training performance we deem
acceptable (following \cite{predictive-maintenance-survey}).
\section{Results and Analysis}
\label{results-and-analysis}

Table~\ref{tab:cost-models} summarizes test-set performance for all models,
averaged across five random seeds. Each model was evaluated on 2,000 held-out machines
containing 66 deterministic failures, with a baseline reactive maintenance
cost of 1,650,000 USD.

\begin{table}[!ht]
  \centering
  \resizebox{\textwidth}{!}{%
  \begin{tabular}{lllllllllll}
    \hline
    \textbf{Model}        & \textbf{Total Cost [USD]} & \textbf{Savings [USD]} & \textbf{Savings \%} & \textbf{Recall} & \textbf{Precision} & \textbf{F1 Score} & \textbf{TP} & \textbf{FP} & \textbf{FN} \\
    \hline
    L0: No-Skill          & 1{,}650{,}000 & 0             & 0.0\%  & 0.0\%  & 0.0\%  & 0.000 & 0  & 0   & 66 \\
    L1: Balanced LogReg   & 796{,}600     & 853{,}400     & 51.7\% & 77.3\% & 13.3\% & 0.227 & 51 & 333 & 15 \\
    L2: Cost-Aware LogReg & 813{,}000     & 837{,}000     & 50.7\% & 82.4\% & 9.8\%  & 0.175 & 54 & 502 & 11 \\
    L3: Cost-Aware Tree   & 576{,}900     & 1{,}073{,}100 & 65.0\% & 90.0\% & 23.6\% & 0.364 & 59 & 229 & 6  \\
    L4: Random Forest     & 481{,}500     & 1{,}168{,}500 & 70.8\% & 96.4\% & 23.6\% & 0.379 & 64 & 207 & 2  \\
    L5: AdaBoost          & 531{,}500     & 1{,}118{,}500 & 67.8\% & 95.2\% & 18.9\% & 0.314 & 63 & 275 & 3  \\
    L6: LightGBM          & 510{,}900     & 1{,}139{,}100 & 69.0\% & 91.5\% & 31.2\% & 0.364 & 60 & 138 & 6  \\
    \textbf{L7: Bayesian SCM} & 554{,}900 & 1{,}095{,}100 & 66.4\% & 90.3\% & 26.2\% & 0.395 & 59 & 193 & 6  \\
    \hline
  \end{tabular}}
  \caption{Model performance, costs, and savings, averaged over validation splits. TP, FP, and FN counts are rounded to the nearest integer.}
  \label{tab:cost-models}
\end{table}

These results reveal a clear performance hierarchy:

\begin{itemize}
  \tightlist

  \item \texttt{Model L0} provides zero value, establishing that any real predictive
    model must outperform the trivial solution of doing nothing.\\

  \item \texttt{Models L1} and \texttt{L2}---variations of logistic
    regression---cluster around a 50--52\% cost reduction, saving between
    837,000 USD and 853,400 USD annually.\\

  \item \texttt{Model L3}, leveraging non-linear decision trees, achieves a 65.0\%
    cost reduction with 1,073,100 USD in annual savings.\\

  \item \texttt{Model L4} (Random Forest) achieves the best performance among all
    models with a 70.8\% cost reduction and 1,168,500 USD in annual savings and
    the highest recall (96.4\%), catching 64 out of 66 failures with only 2 missed.\\

  \item \texttt{Model L5} (AdaBoost) achieves a 67.8\% cost reduction with
    1,118,500 USD in annual savings and 95.2\% recall (63 of 66 failures caught).\\

  \item \texttt{Model L6} (LightGBM) achieves a 69.0\% cost reduction with
    1,139,100 USD in annual savings. Though the model predicted considerably fewer
    false positives compared to random forest, its
    lower recall still results in slightly higher costs due to missed failures.\\

  \item \texttt{Model L7} (Bayesian SCM) achieves a 66.4\% cost reduction with
    1,095,100 USD in annual savings and 90.3\% recall. While this places it behind
    the best ensemble models on raw cost metrics, it is the only model capable of
    attributing predicted failures to specific types, as discussed in
    Section~\ref{subsubsec:failure_attribution}.
\end{itemize}

\subsection{Detailed Model Analysis}
\label{detailed-model-analysis}

The no-skill baseline (\texttt{L0}) catches zero failures and incurs the full
1,650,000 USD reactive maintenance cost. This serves as baseline for evaluating
other models. Logistic regression models and their variants (\texttt{L1},
\texttt{L2}) resulted in 47--55\% cost reduction relative to the baseline. Their
suboptimal results are not surprising considering the nonlinear true data
generating process and the linear decision boundary these models use to make
predictions.

The non-linear tree-based correlation-based models (\texttt{L3}, \texttt{L4},
\texttt{L5}, \texttt{L6}) perform substantially better with 65--71\% cost
reduction. Random Forest (\texttt{L4}) had the best performance as it misses only
2 failures with a reasonable number of false positives. It is worth noting that
the underlying data generating process of this dataset is itself a decision tree.
Yet ensemble models, even equipped with a weak learner like AdaBoost, outperform
a single decision tree. This demonstrates the strength of ensemble methods as
they aggregate multiple estimators and can overcome the single tree's tendency to
overfit. On the other hand, a single decision tree has better interpretability since one
can trace the decision path down the tree to infer the failure mode. This property
is lost in ensemble models, in exchange for superior accuracy and variance
reduction.

The causal model, Bayesian SCM (\texttt{L7}), achieves a 66.4\% cost reduction
and the highest recall among non-ensemble models (90.3\%). Compared to the best
of the correlation-based models, the causal model suffers a modest performance
loss in exchange for stronger interpretability. That is, when
the model predicts a failure, it simultaneously attributes that failure to the
most likely causes. This allows the maintenance team to address the specific
failure mechanism identified by the model, which can be a favorable trade-off in
real-world maintenance settings where understanding why a machine will fail is as
important as knowing that it will.

\subsubsection{Failure Attribution}
\label{subsubsec:failure_attribution}

Among the correlation-based models, the decision tree offers the strongest level of interpretability. One may trace the decision path, record the features that drive the classification and infer the failure types. This, however, requires additional effort to perform the backtracing. Other correlation-based models like ensemble methods present a deeper challenge, as the reasoning is masked behind the aggregation of many weak learners. The causal model, with the domain knowledge encoded in the structure equations, has the inherent ability to attribute failure without needing interpretation efforts.

We examine the quality of failure attribution reported by the Bayesian SCM. Recall that the model was trained using only binary machine failure labels without knowing specific failure types (TWF, HDF, PWF, or OSF). Inside the model it uses structural equations to estimate the probability of each failure mode. When the model predicts a failure, we attribute the failure to the type with the greatest probability. We evaluate the attribution results against the ground truth failure type labels available in the dataset. Table~\ref{tab:attribution} reports the number of ground truth failures in the test set (seed 42), the number detected by the model, and the number correctly attributed among those detected.

\begin{table}[!ht]
  \centering
  \begin{tabular}{lllll}
    \hline
    \textbf{Failure Type} & \textbf{Ground Truth} &  \textbf{Detected} & \textbf{Correctly Attributed} \\
    \hline
    TWF & 7 & 3 & 1 \\
    HDF & 25 & 25 & 25 \\
    PWF & 15 & 15 & 15 \\
    OSF & 22 & 22 & 22\\
    \hline
  \end{tabular}
  \caption{Per-type failure attribution performance on the test set (seed 42).
Attribution analysis is reported for a single representative seed as failure
type labels were not recorded in the averaged results.}
  \label{tab:attribution}
\end{table}

For those detected failures, the Bayesian SCM achieves 100\% correct attribution for HDF, PWF, and OSF types. This encouraging result shows that the structural equations successfully captured the physical mechanisms of each failure type, especially considering that the model was never given the failure type labels during training. TWF attribution, however, presents a unique challenge rooted in the data-generating process itself. As described in the dataset documentation \cite{uci-predictive-maintenance}, TWF involves two sequential random checks. Firstly, each machine is assigned a random wear threshold between 200--240 minutes. Secondly, upon crossing that threshold, whether the machine actually fails or receives a tool replacement is determined by another independent random assignment (46 failures out of 120 threshold crossings in the entire dataset). The first random check can in principle be approximated by a probabilistic model. However, the second check is entirely independent of all sensor readings. This introduces irreducible noise which cannot be resolved by any deterministic classification model. As a result, the model achieves lower performance on TWF attribution compared to other failure types. This is expected given the stochasticity in the data-generating process. For the remaining three failure types, which are governed by deterministic threshold rules on observable sensor features, the Bayesian SCM provides accurate attribution.

\subsubsection{Generalization and Overfitting Assessment}
\label{generalization-and-overfitting-assessment}

We performed the multi-seed validation tactic as explained in \ref{validation-strategy}. Table~\ref{tab:train-test-gap} shows
training savings percentage, test savings percentage, and the gap in percentage
points for predictive models.

\begin{table}[!ht]
  \centering
  \begin{tabular}{llll}
    \hline
    \textbf{Model}        & \textbf{Train Savings \%} & \textbf{Test Savings \%} & \textbf{Gap (\%)} \\
    \hline
    L1: Balanced LogReg   & 57.2\% & 51.7\% & 5.44 \\
    L2: Cost-Aware LogReg & 56.0\% & 50.7\% & 5.26 \\
    L3: Cost-Aware Tree   & 69.4\% & 65.0\% & 4.36 \\
    L4: Random Forest     & 73.3\% & 70.8\% & 2.44 \\
    L5: AdaBoost          & 69.6\% & 67.8\% & 1.82 \\
    L6: LightGBM          & 72.6\% & 69.0\% & 3.60 \\
    L7: Bayesian SCM      & 69.5\% & 66.4\% & 3.17 \\
    \hline
  \end{tabular}
  \caption{Train–test savings comparison averaged over validation splits. Gap represents the difference between
  training and test savings percentages; smaller gaps indicate better
  generalization.}
  \label{tab:train-test-gap}
\end{table}

Most models exhibit modest generalization gaps below 5 percentage points, which
we set as an acceptable bound because plant operators typically budget that
amount of degradation when deploying predictive maintenance models across
production lines \cite{predictive-maintenance-survey}. Notably, logistic
regression variants (\texttt{L1}, \texttt{L2}) are exceptions which marginally
exceed the acceptable bound at 5.44 and 5.26 percentage points respectively.
This failure to generalize is not surprising given the inconsistency between
their linear decision boundary and the nonlinear ground truth. The remaining
models all fall within the acceptable range. AdaBoost (\texttt{L5}) records the
smallest gap at 1.82 percentage points, and Random Forest (\texttt{L4}) follows
at 2.44 points. The Bayesian SCM (\texttt{L7}) achieves a 3.17 point gap,
consistent with the expectation that causal relationships are less susceptible to
distributional idiosyncrasies in the training sample. The multi-seed validation
confirmed stable model rankings across all five splits, demonstrating that
observed performance differences reflect genuine model capabilities rather than
chance variation in data partitioning.

\section{Conclusion}
\label{conclusion}

This study evaluated eight predictive maintenance models on a CNC machine failure
dataset, comparing traditional correlation-based machine learning approaches
with Bayesian structural causal methods. Among correlation-based models, ensemble
methods consistently achieve the best performance, with Random Forest (\texttt{L4})
achieving the best cost reduction of 70.8\%. The Bayesian Structural Causal Model
(\texttt{L7}) achieved competitive financial performance of a 66.4\% cost
reduction. On the other hand, it is capable of attributing predicted failures to
specific types without requiring failure-type labels during training. This capability can prove to be a favorable
trade-off in predictive maintenance scenarios.

This work demonstrates several important findings for predictive maintenance
applications:

\begin{enumerate}
  \tightlist

  \item \textbf{Business-aligned optimization}: Models optimized explicitly for
    business costs (accounting for the 50:1 cost asymmetry between false
    negatives and false positives) substantially outperform accuracy-optimized
    approaches.

  \item \textbf{Ensemble methods as performance ceiling}: Tree-based ensemble
    models (\texttt{L4}, \texttt{L5}, \texttt{L6}) consistently outperform single
    models including the decision tree that mirrors the true data-generating
    process. This demonstrates the strength of ensembling methods in reducing
    variance and preventing overfitting.

  \item \textbf{Causal AI for interpretability}: The Bayesian SCM (\texttt{L7})
    delivers competitive cost savings and provides actionable insights in the form
    of failure attribution. The performance gap relative to the best ensemble model
    is modest and may be outweighed by the practical value of the failure diagnosis.

  \item \textbf{Failure attribution without attribution labels}: The Bayesian SCM
    (\texttt{L7}) correctly attributes predicted failures to specific physical
    mechanisms (TWF, HDF, PWF, OSF) despite being trained only on binary failure
    labels. This means there is no additional labeling cost required for training
    the model.

  \item \textbf{Robust generalization}: The Bayesian SCM demonstrates solid
    generalization (3.17 percentage point train-test gap), which is consistent with
    the expectation that causal relationships are invariant properties of the
    physical system and less susceptible to distributional variation in training
    data.

\end{enumerate}

Several limitations of this study should be acknowledged:

\begin{enumerate}
  \tightlist

  \item \textbf{Dataset scope}: The analysis uses a single dataset of 10,000 CNC
    machines. Generalization to other machine types, manufacturing environments,
    or operational conditions requires additional validation.

  \item \textbf{Causal graph construction}: The causal DAG was manually constructed
    based on domain knowledge of CNC machine failure mechanisms. In domains
    where such expertise is unavailable, causal discovery algorithms could be
    explored, though they introduce additional challenges.

  \item \textbf{Cost estimates}: The business costs (25,000 USD for failures,
    5,000 USD for preventive maintenance, 500 USD for false alarms) are
    representative estimates. Actual costs vary by industry, machine criticality,
    and operational context.

  \item \textbf{Static analysis}: This study evaluates models on synthetic data
    without real-world deployment. Practical implementation would require
    monitoring for distribution drift, model degradation, and integration with
    existing maintenance management systems.

  \item \textbf{Class imbalance}: The 3.3\% failure rate, while realistic for
    well-maintained equipment, limits the number of positive examples available
    for model training. Larger datasets with more failure observations could
    improve model performance.

\end{enumerate}

Future research directions include: (1) validation on operational data;
(2) investigation of causal discovery algorithms to automate DAG construction;
(3) development of online learning approaches for continual model adaptation;
(4) integration with IoT sensor networks and SCADA systems for real-time
deployment; and (5) extension to multi-component systems with cascading failure
modes.

\section{Author Contributions}
The authors are listed in alphabetical order.

Krishna Taduri performed exploratory data analysis, developed the causal structural model (including failure attribution) and integrated cost optimization.

Shaunak Dhande performed exploratory data analysis and developed the correlation-based models.

Chutian Ma reviewed and revised the manuscript, and contributed to the discussion of the failure attribution mechanism.

Giacinto Paolo Saggese and Paul Smith provided overall conceptual guidance, contributed to architectural design decisions, and offered feedback on the manuscript.

\bibliographystyle{amsrn}
\bibliography{references}

\end{document}

%% file: template/style.tex
\documentclass[11pt, reqno]{amsart}
\usepackage{amsfonts, amssymb, amscd, amsrefs}
\usepackage{graphicx}
\usepackage{hyperref}
\usepackage{slashed}
\usepackage{fullpage}
\usepackage{float}
\usepackage{enumitem}
\usepackage[table]{xcolor}
\usepackage{colortbl}
\usepackage{bbold}
\usepackage{placeins}
\usepackage{afterpage}

\usepackage{multirow}
\usepackage{booktabs}
\usepackage{tabularx}
\usepackage{longtable, array}

\usepackage{algorithm}
\usepackage{algpseudocode}

\usepackage{makecell}

\usepackage{amsmath, amssymb, amsthm}
\usepackage{tikz}
\usetikzlibrary{
  positioning,
  calc,
  arrows.meta,
  decorations.pathreplacing,
  shapes.geometric,
  fit
}

\tikzset{
arrow/.style={->, >=stealth, thick},
every picture/.style={ transform shape, scale=0.85, every node/.style={ scale=0.85, inner sep=2pt, outer sep=1pt } },
box/.style={ rectangle, draw, minimum width=3cm, minimum height=1cm, text centered, align=center },
every text node part/.style={ font=\sffamily } }

\usepackage{pgf}
\usepackage[pdf]{graphviz}
\usepackage{uniquecounter}

\definecolor{codebg}{RGB}{245, 245, 230} 

\usepackage[useregional]{datetime2}
\DTMlangsetup[en-US]{zone=eastern,mapzone}

\theoremstyle{plain}

\theoremstyle{definition}

\theoremstyle{remark}

\usepackage{listings}

\newcommand{\pythonstyle}{\lstset{ language=Python, basicstyle=\ttm, morekeywords={self}, 
keywordstyle=\ttb
\color{deepblue}
, emph={MyClass,__init__}, 
emphstyle=\ttb
\color{deepred}
, 
stringstyle=
\color{deepgreen}
, frame=tb, 
showstringspaces=false }}

\lstnewenvironment{python}[1][]{ \pythonstyle \lstset{#1} }{}

\newcommand{\pythoninline}[1]{{\pythonstyle\lstinline!#1!}}

\usepackage{listings}
\usepackage{xcolor}

\definecolor{codegreen}{rgb}{0,0.6,0}
\definecolor{codegray}{rgb}{0.5,0.5,0.5}
\definecolor{codepurple}{rgb}{0.58,0,0.82}
\definecolor{backcolour}{rgb}{0.95,0.95,0.92}
\definecolor{redbackground}{rgb}{1,0.9,0.9}
\definecolor{greenbackground}{rgb}{0.9,1,0.9}

\lstdefinestyle{mystyle}{ backgroundcolor=
\color{backcolour}
, commentstyle=
\color{codegreen}
, keywordstyle=
\color{magenta}
, numberstyle=\tiny
\color{codegray}
, stringstyle=
\color{codepurple}
, basicstyle=\ttfamily\footnotesize, breakatwhitespace=false, breaklines=true, captionpos=b,
keepspaces=true, numbers=left, numbersep=5pt, showspaces=false, showstringspaces=false,
showtabs=false, tabsize=2 }

\lstdefinestyle{redstyle}{ backgroundcolor=
\color{redbackground}
, commentstyle=
\color{codegreen}
, keywordstyle=
\color{magenta}
, numberstyle=\tiny
\color{codegray}
, stringstyle=
\color{codepurple}
, basicstyle=\ttfamily\small, breakatwhitespace=false, breaklines=true, captionpos=b,
keepspaces=true, showspaces=false, showstringspaces=false,
showtabs=false, tabsize=2 }

\lstdefinestyle{greenstyle}{ backgroundcolor=
\color{greenbackground}
, commentstyle=
\color{codegreen}
, keywordstyle=
\color{magenta}
, numberstyle=\tiny
\color{codegray}
, stringstyle=
\color{codepurple}
, basicstyle=\ttfamily\small, breakatwhitespace=false, breaklines=true, captionpos=b,
keepspaces=true, showspaces=false, showstringspaces=false,
showtabs=false, tabsize=2 }

\lstdefinestyle{codebgstyle}{ backgroundcolor=
\color{codebg}
, commentstyle=
\color{codegreen}
, keywordstyle=
\color{magenta}
, numberstyle=\tiny
\color{codegray}
, stringstyle=
\color{codepurple}
, basicstyle=\ttfamily\small, breakatwhitespace=false, breaklines=true, captionpos=b,
keepspaces=true, numbers=left, numbersep=5pt, showspaces=false, showstringspaces=false,
showtabs=false, tabsize=2 }

\providecommand{\tightlist}{%
\setlength{\itemsep}{0pt}
\setlength{\parskip}{0pt}}

\makeatletter
\newcommand{\printfnsymbol}[1]{%
\textsuperscript{\@fnsymbol{#1}}%
}
\makeatother